\documentclass{article} 
\usepackage{iclr2027_conference_arxiv,times}
\iclrarxivcopy

\usepackage{amsmath,amsfonts,bm}

\def\eqref#1{equation~\ref{#1}}

\def\1{\bm{1}}

\DeclareMathAlphabet{\mathsfit}{\encodingdefault}{\sfdefault}{m}{sl}
\SetMathAlphabet{\mathsfit}{bold}{\encodingdefault}{\sfdefault}{bx}{n}

\usepackage{hyperref}
\usepackage{url}
\usepackage{amsmath}
\usepackage{amssymb}
\usepackage{algorithm}
\usepackage{algpseudocode}
\usepackage{booktabs}
\usepackage{graphicx}
\usepackage[table]{xcolor}
\usepackage{enumitem}
\usepackage{caption}
\usepackage{siunitx}

\definecolor{qwenbg}{HTML}{DDEBF7}
\definecolor{dopdbg}{HTML}{E2F0D9}

\title{d-OPD: Future-Aware On-Policy Distillation for Block Diffusion Language Models}

\begin{document}

\author{%
\parbox{\linewidth}{\centering
{\bfseries
Ruitao Liu\textsuperscript{1}
\quad
Qinghao Hu\textsuperscript{2}
\quad
Song Han\textsuperscript{2,3}
}
\\[1.0ex]
{\normalfont
\textsuperscript{1}Tsinghua University \quad
\textsuperscript{2}MIT \quad
\textsuperscript{3}NVIDIA
}
}}
\maketitle

\begin{abstract}
Large language models (LLMs) typically generate text autoregressively (AR), predicting one token at a time. Block diffusion language models (dLLMs) instead generate blocks sequentially while denoising multiple tokens in parallel within each block, offering a promising way to accelerate generation. Rather than training such models from scratch, recent work adapts strong pretrained AR models into block dLLMs through distillation. On-policy distillation (OPD) has been widely used for LLM training because it supervises the student on states generated by its current policy, rather than only on fixed offline trajectories. By training on the states the student actually visits, it reduces the mismatch between training and generation and can provide more relevant supervision as the student evolves. Recent work has extended this idea to AR-to-block-diffusion conversion. However, this setting introduces a fundamental mismatch in supervision: the block-diffusion student and the causal AR teacher condition on different information at the same training state. The student predicts from the entire partially denoised block, including visible future context, whereas the standard AR teacher target is defined only from the causal prefix. As a result, the teacher distribution used for distillation is not fully aligned with the information available to the student. We therefore introduce d-OPD, a future-aware on-policy distillation method that corrects the AR teacher distribution to better align with the student-visible state by incorporating visible future information within each block, providing supervision that better matches the information used by the student. Across Qwen3 models from 0.6B to 8B, d-OPD improves the six-benchmark average by up to $4.0$ points over OPDLM and reduces training time by $1.35$--$1.58\times$.
The code is available at \url{https://github.com/mit-han-lab/d-OPD}.
\end{abstract}

\section{Introduction}

Autoregressive (AR) language modeling is the standard approach used by most modern large language models (LLMs)~\citep{SequenceLevel2025ACL,CharacterDLMAR2025Arxiv,ReasoningBias2025Arxiv,AudioLM2025ICML,DiscreteDiff2025Arxiv,AREnergy2026ICML,SDAR2026ACL,RLP2026ICLR,PlannedDiff2026ICLR,ReFusion2026ICLR,KForcing2026Arxiv}. AR models generate one token at a time, with each new token depending on the tokens generated before it. This token-by-token dependency makes inference slow, especially for long outputs. To reduce this sequential bottleneck, diffusion language models (dLLMs) generate multiple token positions in parallel for more efficient inference~\citep{Learn2Parallel2026ICLR,dParallel2026ICLR,Bits2Rounds2026ICML,FlashDLM2026ICLR,AccelerateDiff2025Arxiv,Mercury2025Arxiv,FastFluentDiff2025NIPS,FastDLLMV22026ICLR,Dream2025Arxiv,ExperimentalDiff2026Arxiv,DMax2026Arxiv,DiffSurvey2026Arxiv,SoftMaskDiff2026ICLR,DiffStruggle2026ICLR,CLAD2026Arxiv,d3LLM2026ICML,DiRL2026Arxiv}. Among dLLMs, block diffusion has become a particularly important design because it can better preserve the generation quality of AR models while retaining parallel generation within each block~\citep{BlockDiff2025ICLR,LLaDA2025Arxiv,SDAR-VL2026ACL,SDLM2025Arxiv,SDAR2026ACL,PARD2026Arxiv,MultiBlock2026Arxiv,AdapBlockDiff2026Arxiv,DiffInDiff2026Arxiv,MultiBLockEditing2026Arxiv,CacheBlockDiff2026Arxiv}. Block dLLMs generate blocks from left to right, preserving the causal structure across blocks while allowing tokens within the current block to interact bidirectionally.

Rather than training such models from scratch, a particularly attractive direction is to convert existing pretrained AR checkpoints into block dLLMs~\citep{LLaDA2025Arxiv,SDLM2025Arxiv,FastDLLMV22026ICLR,SDAR2026ACL,EfficientDLM2026Arxiv,FLUID2026ACL,NextToken2NextBlock2026Arxiv,BlockVLA2026Arxiv,DiffusionVL2026Arxiv}. As strong pretrained AR models contain broad linguistic and task knowledge learned through large-scale pretraining, conversion methods aim to preserve this knowledge while adapting the model to block-wise parallel generation. Distillation provides a natural way to achieve this by using the AR model as a teacher during conversion. However, standard offline distillation supervises the student on states from fixed offline trajectories rather than the partially denoised states it actually visits during block-dLLM generation. On-policy distillation (OPD) addresses this state-distribution shift by training on states generated by the student's current policy~\citep{dOPSD2026Arxiv,RethinkOPD22026Arxiv,OPDeltaD2026Arxiv,GuidedOPSD2026Arxiv,RethinkOPD2026ICML,UOPSD2026Arxiv,OPDSurvey2026Arxiv,SSOPSD2026Arxiv,OPRD2026Arxiv,OPSDL2026Arxiv,OPSD2026ICML}. OPDLM applies this idea to block-dLLM conversion: it rolls out the block-dLLM student on-policy and uses an AR teacher to provide token-level supervision on the resulting student-generated states~\citep{OPDLM2026Arxiv}.

However, matching the state distribution is not sufficient for block-dLLM distillation. A fundamental mismatch remains in supervision because the block-dLLM student and the causal AR teacher condition on different information at the same training state. Consider a masked position inside a partially denoised block. The student predicts this token from the entire visible block state, which may already contain visible tokens to its right. The standard AR teacher target, in contrast, is defined only from the causal prefix. Consequently, even when teacher supervision is queried on exactly the state visited by the student, the resulting teacher distribution does not fully reflect the information available to the student. OPD therefore only aligns the student’s training and generation states, while leaving the teacher target misaligned with the student’s available information.

This mismatch suggests that the target distribution should reflect the visible future information inside the current block. The key question is how the teacher distribution at the current position should change given this future context. By the chain rule, the AR teacher's next-token probabilities can be combined to compute the joint probability of a complete sequence. This joint distribution actually allows visible future tokens to provide evidence about the current masked token. This is because different candidate values at the current position can make the observed future more or less likely, which in turn provides a way to reweight the teacher distribution at the current position.

Using this joint-distribution view, we introduce \textbf{d-OPD}, a future-aware OPD method for AR-to-block-dLLM conversion. d-OPD uses the visible future context to correct the AR teacher distribution at each masked position. Specifically, we formalize the desired target as the AR teacher posterior conditioned on the complete student-visible state, which naturally combines the causal teacher prediction with how well each candidate agrees with the visible future. Because exact conditioning requires intractable marginalization over undenoised prefix and future tokens, d-OPD constructs a tractable approximation from completed on-policy trajectories. This approximation provides future-aware teacher supervision that better matches the information available to the block-dLLM student.

We evaluate d-OPD across Qwen3 models from 0.6B to 8B. d-OPD consistently improves over OPDLM across model scales and block sizes, with gains of $2.6$--$4.0$ points in the six-benchmark average. The improvement remains stable as the model scales from 0.6B to 8B, showing that the benefit is not limited to a particular model size. It also persists as the block size increases, reaching the largest improvement of $4.0$ points at $N=8$, where more token positions are modeled jointly within each block. Beyond accuracy, d-OPD improves training efficiency as well, reducing the wall-clock time required to reach the corresponding OPDLM performance levels by $1.35$--$1.58\times$ across all evaluated model scales and training settings.

We summarize our contributions as follows:
\begin{itemize}[leftmargin=1.6em,itemsep=0.4em,topsep=0.0em,parsep=0pt]
    \item We identify a fundamental \emph{target-conditioning mismatch} in AR-to-dLLM OPD. Although OPD aligns the student states used for training with those encountered during generation, the causal AR teacher and block-dLLM student still condition on different information.
    \item We introduce \textbf{d-OPD}, a future-aware OPD method that corrects the AR teacher distribution using visible future information within each block. We formalize the desired future-conditioned target and develop a tractable trajectory-based correction for standard OPD training.
    \item We evaluate d-OPD across Qwen3 models from 0.6B to 8B. d-OPD consistently improves over OPDLM, achieving up to $4.0$ points higher six-benchmark average and $1.35$--$1.58\times$ faster time to the corresponding OPDLM performance across the evaluated settings.
\end{itemize}
\begin{figure*}[t]
    \centering
    \includegraphics[width=\textwidth]{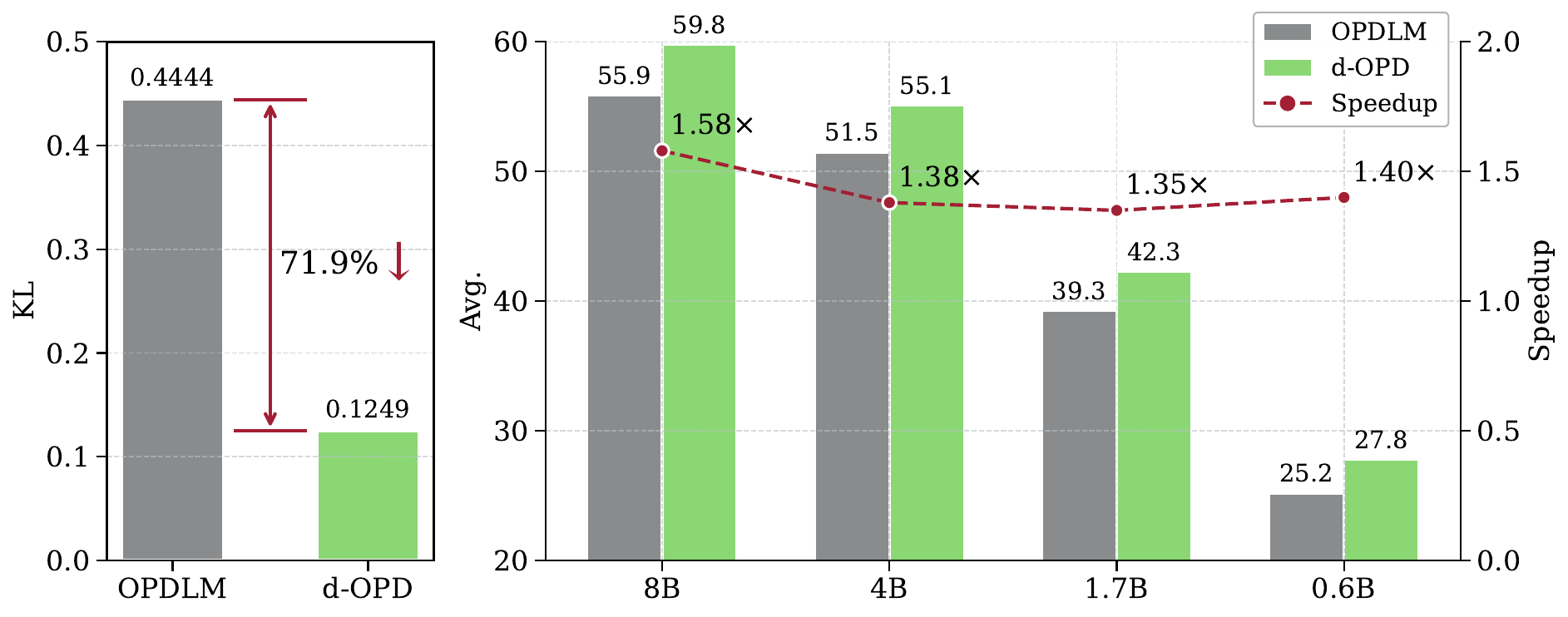}
    \caption{\textbf{d-OPD results.} \textbf{Left:} Correcting the AR teacher with visible future information substantially reduces its mismatch with the exact student-state-conditioned target, lowering the mean token-level KL by 71.9\%. \textbf{Right:} d-OPD consistently improves the six-benchmark average over OPDLM across Qwen3 models from 0.6B to 8B, with training speedup of up to $1.58\times$.}
    \label{fig:teaser}
\end{figure*}
\section{Related Work}
\label{sec:background}

\subsection{Diffusion language models}
Diffusion language models (dLLMs) generate text by iteratively denoising masked tokens and have recently scaled to general language modeling, reasoning, and instruction following~\citep{Learn2Parallel2026ICLR,dParallel2026ICLR,Bits2Rounds2026ICML,FlashDLM2026ICLR,AccelerateDiff2025Arxiv,Mercury2025Arxiv,FastFluentDiff2025NIPS,FastDLLMV22026ICLR,Dream2025Arxiv,ExperimentalDiff2026Arxiv,DMax2026Arxiv,DiffSurvey2026Arxiv,SoftMaskDiff2026ICLR,DiffStruggle2026ICLR,CLAD2026Arxiv,d3LLM2026ICML,DiRL2026Arxiv}. Block dLLMs further divide generation into causal blocks while allowing bidirectional interaction within the current block~\citep{BlockDiff2025ICLR,LLaDA2025Arxiv,BlockVLA2026Arxiv}. Recent work improves block dLLM inference through confidence-aware decoding, cache reuse, alternative block schedules, generation orders, and specialized serving systems~\citep{FastDLLM2026ICLR,PARD2026Arxiv,Sangam2026Arxiv,FlowBlock2026Arxiv,GroupKV2026ICPP}. These works mainly improve inference for trained dLLMs, whereas we study the teacher target used during AR-to-block-dLLM conversion.

\subsection{AR-to-diffusion adaptation}
Recent work converts pretrained AR models into dLLMs through changes in attention, masking, denoising, and block-wise training while preserving pretrained knowledge~\citep{LLaDA2025Arxiv,ScaleDiff2025ICLR,SDLM2025Arxiv,FLUID2026ACL,FastDLLMV22026ICLR,Dream2025Arxiv,SDAR2026ACL,EfficientDLM2026Arxiv,NextToken2NextBlock2026Arxiv,BlockVLA2026Arxiv,DiffusionVL2026Arxiv}. BARD, T$^\star$, and related methods refine this conversion through progressive block scaling, stage-wise distillation, trajectory-aware curricula, auxiliary AR objectives, and on-policy adaptation~\citep{LLaDA2025Arxiv,BARD-VL2026Arxiv,OPDLM2026Arxiv,NextToken2NextBlock2026Arxiv,TStar2026ACL,FastDVLM2026Arxiv}.

\subsection{On-policy distillation}
Knowledge distillation transfers teacher behavior through token- or sequence-level supervision~\citep{KDSurvey2025TMLR,KDLM2025ACL,KDLLMSurvey2025ACM,KDDD2026Arxiv}. On-policy distillation (OPD) trains on student-generated states, reducing the mismatch between training and generation~\citep{RethinkOPD22026Arxiv,RethinkOPD2026ICML,OPDSurvey2026Arxiv}. Recent work has extensively studied and extended OPD for LLM post-training~\citep{dOPSD2026Arxiv,RethinkOPD22026Arxiv,OPDeltaD2026Arxiv,GuidedOPSD2026Arxiv,RethinkOPD2026ICML,UOPSD2026Arxiv,OPDSurvey2026Arxiv,SSOPSD2026Arxiv,OPRD2026Arxiv,OPSDL2026Arxiv,OPSD2026ICML}. OPDLM applies this paradigm to AR-to-dLLM conversion~\citep{OPDLM2026Arxiv}. It aligns the student states used for training with those encountered during generation, but leaves a target-conditioning mismatch between the causal teacher and block-dLLM student.
\section{Motivation}
\label{sec:motivation}

\subsection{Target-conditioning mismatch}

Consider a block dLLM with block size $N$. Let $q$ denote the input question and let $z$ denote the completed causal context preceding the active block. We denote the active block by
\begin{equation}
X=(x_1,\ldots,x_N).
\end{equation}

At denoising step $t$, the block dLLM student visits a partially masked state
\begin{equation}
X_t=(x_{t,1},\ldots,x_{t,N}).
\end{equation}
Let $p_{\theta,i}$ denote the student distribution at position $i$, where $\theta$ denotes the student parameters. Some entries of $X_t$ have already been revealed while the rest remain masked. For a masked position $i$, the student predicts from all currently visible tokens in the block,
\begin{equation}
p_{\theta,i}(\cdot\mid X_t,q,z),
\end{equation}
so visible tokens on either side of position $i$ can influence the prediction.

Let $p_T$ denote the distribution defined by the frozen AR teacher. Let
\begin{equation}
\hat X=(\hat x_1,\ldots,\hat x_N)
\end{equation}
denote the completed, fully unmasked block produced by the student rollout. Let $\mathcal V$ denote the vocabulary. For a candidate token $v\in\mathcal V$, standard AR supervision uses the completed causal prefix and defines
\begin{equation}
r_i(v)=p_T(v\mid q,z,\hat x_{<i}).
\end{equation}
Here $r_i$ denotes the causal teacher target at position $i$. This target depends only on the causal prefix and therefore ignores any visible information to the right of $i$. If $x_i$ is masked while tokens to its right are already visible, the block dLLM student can use this future context directly, whereas the causal AR target cannot condition on it. The teacher target and student prediction therefore depend on different information at the same student-visited state, creating a target-conditioning mismatch.

\subsection{Exact future-conditioned target}

To derive the target that resolves this mismatch, we start from the teacher's joint distribution over the active block. The frozen AR teacher defines
\begin{equation}
p_T(X\mid q,z)=\prod_{i=1}^{N}p_T(x_i\mid q,z,x_{<i}).
\label{eq:ar-factorization}
\end{equation}
Under this joint distribution, each candidate value for $x_i$ assigns a likelihood to the observed future context. Different candidates can assign different likelihoods to the same observed future. By Bayes' rule, the causal teacher distribution at position $i$ is reweighted by these likelihoods, yielding the teacher posterior conditioned on the complete student-visible state. 

We first define this posterior at the block level. For a clean block $X$, write $X\succeq X_t$ if $X$ agrees with every visible token in $X_t$. Conditioning the teacher joint distribution on this consistency event gives
\begin{equation}
\pi(X\mid X_t,q,z)=
\frac{p_T(X\mid q,z)\mathbf{1}\{X\succeq X_t\}}
{\sum_{X'}p_T(X'\mid q,z)\mathbf{1}\{X'\succeq X_t\}}.
\label{eq:block-posterior}
\end{equation}
Here $X'$ ranges over all fully unmasked realizations of the active block, and $\mathbf{1}\{\cdot\}$ denotes the indicator function. 

We then define the exact teacher target for a masked position $i$ as the corresponding token-level marginal:
\begin{equation}
\pi_i(v\mid X_t,q,z)=
\sum_{X:x_i=v}\pi(X\mid X_t,q,z).
\label{eq:token-posterior}
\end{equation}
Here $\pi_i$ denotes the corresponding token-level posterior at position $i$.

To expose the structure of this target, let $\mathcal{A}_i(X_t)$ denote the set of left-prefix completions $A_i=x_{<i}$ compatible with $X_t$. Then
\begin{equation}
\pi_i(v\mid X_t,q,z)=
\sum_{A_i\in\mathcal{A}_i(X_t)}
\pi(A_i\mid X_t,q,z)\,
\pi_i(v\mid X_t,q,z,A_i).
\label{eq:left-marginal}
\end{equation}

Now fix one compatible prefix $A_i$. Let $O_i$ denote the visible tokens to the right of position $i$. Applying Bayes' rule to the AR joint distribution gives
\begin{equation}
\pi_i(v\mid X_t,q,z,A_i)
=
\frac{
p_T(v\mid q,z,A_i)\,
p_T(O_i\mid q,z,A_i,v)
}{
\sum_{v'\in\mathcal V}
p_T(v'\mid q,z,A_i)\,
p_T(O_i\mid q,z,A_i,v')
}.
\label{eq:future-factorization}
\end{equation}

Equation~\ref{eq:future-factorization} gives the key observation behind d-OPD. We refer to the first factor, $p_T(v\mid q,z,A_i)$, as the \emph{causal teacher prior}, since it is the ordinary AR teacher distribution over candidate token $v$. We refer to the second factor, $p_T(O_i\mid q,z,A_i,v)$, as the \emph{future-compatibility term}, since it measures how likely the visible future is under that candidate. Future-aware supervision can therefore be obtained by correcting the causal teacher distribution according to the future-compatibility term.

If no future token is visible, then $O_i=\varnothing$ and the future-compatibility term is constant in $v$, so the target reduces to the causal teacher prior. If the entire left prefix is visible, $\mathcal{A}_i(X_t)$ contains only one completion and the outer marginalization disappears. In general, the exact target must account for uncertainty in both the masked prefix and the unobserved future.

\begin{figure*}[t]
    \centering
    \includegraphics[width=\textwidth]{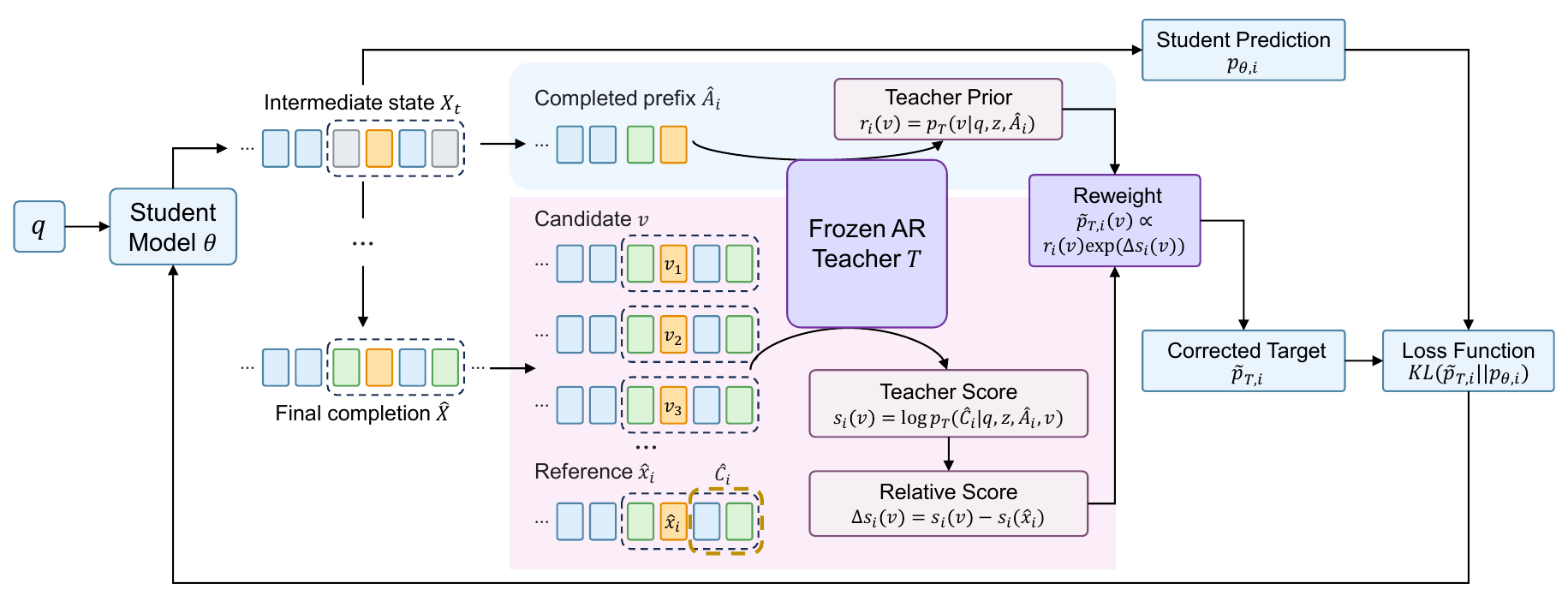}
\captionsetup{font={stretch=1.08}}
    \caption{\textbf{Method overview.} d-OPD starts from a student-visited partially masked block $X_t$ and its completed block $\hat X$ from the same on-policy rollout. For each masked position $i$, the completed prefix $\hat A_i$ gives the causal AR teacher prior $r_i$. A candidate token $v$ and the reference token $\hat x_i$ are scored by how likely the completed future $\hat C_i$ is under the frozen teacher. Their relative future scores are then used to reweight $r_i$ to form the future-aware teacher target $\tilde p_{T,i}$. The block dLLM student is then trained by minimizing the KL divergence between $\tilde p_{T,i}$ and $p_{\theta,i}$.}

    \label{fig:main}
\end{figure*}

\subsection{Diagnosis}
\label{sec:diagnosis}

We quantify the target-conditioning mismatch over 49,948 masked-token marginals in a controlled setting described in Appendix~\ref{app:enumerable}, where the exact future-conditioned posterior can be computed by enumeration. This setting allows us to directly evaluate how closely different teacher targets match the posterior induced by the same partially observed block state, providing a target-level diagnosis of the mismatch rather than inferring it only indirectly from downstream task performance alone. The causal target has a mean token-level KL of $0.4444$ to the exact posterior, compared with $0.1249$ using our method introduced in Section~\ref{sec:method}, a $71.9\%$ reduction. The large gap confirms that conditioning only on the causal prefix can substantially misalign the teacher target with the information available to the block dLLM student. Incorporating future information closes most of this gap, moving the teacher target much closer to the exact posterior. This controlled diagnosis therefore provides direct empirical evidence for the target-conditioning mismatch and motivates explicitly correcting the teacher distribution using student-visible future context.

\section{Method}
\label{sec:method}

\subsection{Future-aware correction}

Section~\ref{sec:motivation} defines the exact future-conditioned teacher target that matches the information available to the block dLLM student. Computing this target is intractable in practice because it requires marginalizing over all compatible left-prefix completions and unobserved future tokens.

d-OPD builds on the completed block $\hat X$ already produced by the on-policy rollout. For a masked position $i$, let $\hat A_i=\hat x_{<i}$ denote the completed prefix from the same rollout. We use $\hat A_i$ as one compatible realization of the uncertain prefix $A$ in Equation~\ref{eq:left-marginal}. The causal teacher prior is therefore
\begin{equation}
r_i(v)=p_T(v\mid q,z,\hat A_i).
\label{eq:causal-prior}
\end{equation}
The corresponding fixed-prefix form of Equation~\ref{eq:future-factorization} is
\begin{equation}
\pi_i(v\mid X_t,q,z,\hat A_i)
=
\frac{
r_i(v)\,
p_T(O_i\mid q,z,\hat A_i,v)
}{
\sum_{v'\in\mathcal V}
r_i(v')\,
p_T(O_i\mid q,z,\hat A_i,v')
}.
\label{eq:fixed-prefix-target}
\end{equation}
Thus the prefix uncertainty has been resolved using $\hat A_i$, and the remaining challenge is to make the future-compatibility term in Equation~\ref{eq:fixed-prefix-target} tractable.

The completed block also provides one compatible realization of the uncertain future. Let
\begin{equation}
\hat C_i=(\hat x_{i+1},\ldots,\hat x_N)
\label{eq:completed-suffix}
\end{equation}
denote the completed suffix after position $i$. We use the likelihood of this completed future under each candidate token \(v\) to estimate the future-compatibility term in Equation~\ref{eq:fixed-prefix-target}:
\begin{equation}
s_i(v)=\log p_T(\hat C_i\mid q,z,\hat A_i,v).
\label{eq:suffix-score}
\end{equation}
Using this completed-future likelihood in Equation~\ref{eq:fixed-prefix-target} gives the completed-future target
\begin{equation}
\tilde p_{T,i}(v)
=
\frac{
r_i(v)\exp(s_i(v))
}{
\sum_{v'\in\mathcal V}
r_i(v')\exp(s_i(v'))
}.
\label{eq:completed-future-target}
\end{equation}
Thus, candidates that make the completed future more likely receive larger teacher weight.

Evaluating $s_i(v)$ for every token in the vocabulary is computationally expensive, so in practice we compute future scores only for a small set of high-probability candidates. This leaves the remaining tokens without a future score. Simply leaving these unscored tokens unchanged while multiplying the scored candidates by their absolute future likelihoods $\exp(s_i(v))$ would be inappropriate, because an absolute sequence likelihood is typically much smaller than one, whereas an unchanged token implicitly receives a correction factor of one. This would artificially favor unscored tokens and distort their relative probability mass. We therefore normalize the future likelihoods by a reference candidate, so that a correction factor of one represents a meaningful neutral baseline. Since $\hat x_i$ is the token actually produced by the rollout, it provides a natural reference. We express each scored likelihood relative to $\hat x_i$:
\begin{equation}
\Delta s_i(v)=s_i(v)-s_i(\hat x_i).
\label{eq:relative-score}
\end{equation}
Equivalently,
\begin{equation}
\exp\!\left(\Delta s_i(v)\right)
=
\frac{
p_T(\hat C_i\mid q,z,\hat A_i,v)
}{
p_T(\hat C_i\mid q,z,\hat A_i,\hat x_i)
}.
\label{eq:relative-future-ratio}
\end{equation}
Dividing by the reference likelihood does not change the full-vocabulary target, since the same factor cancels during normalization:
\begin{equation}
\tilde p_{T,i}(v)
=
\frac{
r_i(v)\exp(\Delta s_i(v))
}{
\sum_{v'\in\mathcal V}
r_i(v')\exp(\Delta s_i(v'))
}.
\label{eq:relative-future-target}
\end{equation}
For a candidate set $V_i$, we apply the relative future correction only to scored candidates in $V_i$, while tokens outside $V_i$ are left at the baseline correction weight one.

\subsection{Method details}
For future scoring, we use the union of the teacher and student top-$k$ sets as the candidate set, since high-probability tokens capture most of the probability mass under both distributions. Taking the union ensures that candidates considered important by either distribution can receive future-aware correction, while avoiding full-vocabulary continuation scoring. As shown in Table~\ref{tab:ablation}, the top-$4$ tokens already cover more than $95\%$ of the probability mass for both the teacher and the student, and the coverage quickly saturates as $k$ increases. This allows future-aware correction to focus on only a small set of plausible candidates, while leaving the remaining vocabulary unchanged.

We also use a correctness gate for the future-aware correction: the completed future is used to reweight the teacher target only when the rollout passes the task verifier, since futures from incorrect rollouts may provide noisy or misleading evidence for reweighting and would also incur unnecessary scoring cost, while standard causal teacher supervision is still applied to all rollouts. More implementation details are provided in Algorithm~\ref{alg:training} in Appendix~\ref{app:algorithm}.


\definecolor{qwenbg}{HTML}{DDEBF7}
\definecolor{dopdbg}{HTML}{E2F0D9}

\begin{table}[t]
\centering
\caption{\textbf{Main results.} Accuracy across block sizes and Qwen3 model scales on six benchmarks. d-OPD consistently outperforms OPDLM across all evaluated model scales, with improvements remaining stable from 0.6B to 8B. The gains also persist as the block size increases, reaching up to $4.0$ points in the six-benchmark average. Best results are in bold. Second-best results are underlined.}
\label{tab:main}

\small
\setlength{\tabcolsep}{3.0pt}

\begin{tabular}{@{}llccccccc@{}}
\toprule

Setting & Method
& MMLU & MMLU-P & GPQA-D
& GSM8K & MATH500 & AIME25
& Avg. \\
\midrule

\multicolumn{9}{l}{\textit{Qwen3-8B: comparison across block sizes}} \\
\midrule

&
\cellcolor{qwenbg}Qwen3-8B
& \cellcolor{qwenbg}76.1
& \cellcolor{qwenbg}60.5
& \cellcolor{qwenbg}44.4
& \cellcolor{qwenbg}91.7
& \cellcolor{qwenbg}83.2
& \cellcolor{qwenbg}16.7
& \cellcolor{qwenbg}62.1 \\

$N=4$
& SFT
& 69.1
& 54.3
& \textbf{39.9}
& 87.3
& 75.4
& 13.3
& 56.6 \\

& B-SFT
& \underline{70.9}
& 52.6
& \underline{39.4}
& 88.5
& 72.6
& \underline{16.7}
& 56.8 \\

& BARD
& 68.6
& \underline{55.3}
& 38.4
& \underline{88.6}
& 76.4
& \underline{16.7}
& \underline{57.3} \\

& OPDLM
& 69.2
& 54.0
& 38.4
& 87.2
& \underline{76.8}
& 10.0
& 55.9 \\

&
\cellcolor{dopdbg}\textbf{d-OPD}
& \cellcolor{dopdbg}\textbf{74.3}
& \cellcolor{dopdbg}\textbf{55.9}
& \cellcolor{dopdbg}38.9
& \cellcolor{dopdbg}\textbf{89.3}
& \cellcolor{dopdbg}\textbf{77.2}
& \cellcolor{dopdbg}\textbf{23.3}
& \cellcolor{dopdbg}\textbf{59.8} \\

\cmidrule(lr){1-9}

$N=8$
& OPDLM
& 65.0
& 45.3
& 24.2
& 77.3
& 59.0
& 6.7
& 46.3 \\

&
\cellcolor{dopdbg}\textbf{d-OPD}
& \cellcolor{dopdbg}\textbf{67.7}
& \cellcolor{dopdbg}\textbf{48.0}
& \cellcolor{dopdbg}\textbf{27.8}
& \cellcolor{dopdbg}\textbf{83.6}
& \cellcolor{dopdbg}\textbf{61.2}
& \cellcolor{dopdbg}\textbf{13.3}
& \cellcolor{dopdbg}\textbf{50.3} \\

\cmidrule(lr){1-9}

$N=16$
& OPDLM
& 64.5
& 43.3
& 22.7
& 75.4
& 49.6
& \textbf{3.3}
& 43.1 \\

&
\cellcolor{dopdbg}\textbf{d-OPD}
& \cellcolor{dopdbg}\textbf{67.1}
& \cellcolor{dopdbg}\textbf{46.5}
& \cellcolor{dopdbg}\textbf{29.8}
& \cellcolor{dopdbg}\textbf{80.6}
& \cellcolor{dopdbg}\textbf{53.8}
& \cellcolor{dopdbg}\textbf{3.3}
& \cellcolor{dopdbg}\textbf{46.9} \\

\midrule
\multicolumn{9}{l}{\textit{Qwen3: generalization across model scales}} \\
\midrule

4B
& \cellcolor{qwenbg}Qwen3-4B
& \cellcolor{qwenbg}72.1
& \cellcolor{qwenbg}56.2
& \cellcolor{qwenbg}39.9
& \cellcolor{qwenbg}90.4
& \cellcolor{qwenbg}81.8
& \cellcolor{qwenbg}16.7
& \cellcolor{qwenbg}59.5 \\

& SFT
& 65.5
& 48.9
& 29.8
& 82.3
& 74.0
& 6.7
& 51.2 \\

& B-SFT
& 64.2
& \underline{50.2}
& \underline{31.3}
& \underline{87.4}
& 74.6
& \underline{13.3}
& \underline{53.5} \\

& BARD
& 65.1
& \underline{50.2}
& 29.3
& 82.0
& \underline{75.2}
& 10.0
& 52.0 \\

& OPDLM
& \underline{66.2}
& 47.8
& 30.3
& 82.9
& 75.0
& 6.7
& 51.5 \\

&
\cellcolor{dopdbg}\textbf{d-OPD}
& \cellcolor{dopdbg}\textbf{67.4}
& \cellcolor{dopdbg}\textbf{50.5}
& \cellcolor{dopdbg}\textbf{32.3}
& \cellcolor{dopdbg}\textbf{88.3}
& \cellcolor{dopdbg}\textbf{75.4}
& \cellcolor{dopdbg}\textbf{16.7}
& \cellcolor{dopdbg}\textbf{55.1} \\

\cmidrule(lr){1-9}

1.7B
& \cellcolor{qwenbg}Qwen3-1.7B
& \cellcolor{qwenbg}58.6
& \cellcolor{qwenbg}39.4
& \cellcolor{qwenbg}28.3
& \cellcolor{qwenbg}81.5
& \cellcolor{qwenbg}72.0
& \cellcolor{qwenbg}6.7
& \cellcolor{qwenbg}47.7 \\

& OPDLM
& 42.9
& 29.9
& 22.2
& 73.9
& \textbf{60.2}
& \textbf{6.7}
& 39.3 \\

&
\cellcolor{dopdbg}\textbf{d-OPD}
& \cellcolor{dopdbg}\textbf{55.2}
& \cellcolor{dopdbg}\textbf{30.6}
& \cellcolor{dopdbg}\textbf{27.8}
& \cellcolor{dopdbg}\textbf{74.3}
& \cellcolor{dopdbg}59.4
& \cellcolor{dopdbg}\textbf{6.7}
& \cellcolor{dopdbg}\textbf{42.3} \\

\cmidrule(lr){1-9}

0.6B
& \cellcolor{qwenbg}Qwen3-0.6B
& \cellcolor{qwenbg}46.0
& \cellcolor{qwenbg}27.3
& \cellcolor{qwenbg}24.2
& \cellcolor{qwenbg}63.5
& \cellcolor{qwenbg}52.4
& \cellcolor{qwenbg}6.7
& \cellcolor{qwenbg}36.7 \\

& OPDLM
& 41.1
& 18.8
& 19.7
& 40.6
& 31.0
& \textbf{0.0}
& 25.2 \\

&
\cellcolor{dopdbg}\textbf{d-OPD}
& \cellcolor{dopdbg}\textbf{46.7}
& \cellcolor{dopdbg}\textbf{19.9}
& \cellcolor{dopdbg}\textbf{21.7}
& \cellcolor{dopdbg}\textbf{44.6}
& \cellcolor{dopdbg}\textbf{33.6}
& \cellcolor{dopdbg}\textbf{0.0}
& \cellcolor{dopdbg}\textbf{27.8} \\

\bottomrule
\end{tabular}
\end{table}

\subsection{Training}

For each block, we randomly sample one denoising state $X_t$ from the on-policy rollout. This ensures that the student is trained directly on partially revealed states encountered under its current generation policy, rather than on states constructed from a separate offline distribution. Distillation is applied to all masked positions in the sampled state, with the corresponding teacher targets constructed from the completed on-policy rollout. For each masked position $i$, we optimize the forward KL from the corrected teacher target to the student:
\begin{equation}
\mathcal{L}_i(\theta)=
D_{\mathrm{KL}}\!\left(
\tilde p_{T,i}
\mathrel{\|}
p_{\theta,i}
\right).
\label{eq:training-loss}
\end{equation}
The token-level losses are averaged over all masked positions collected across the sampled denoising states of a rollout, so each masked prediction contributes directly to the distillation objective. Sampling one state per block also allows a single rollout to provide supervision at different denoising states across its blocks. d-OPD changes only the teacher supervision at these positions, while keeping the student architecture, distillation objective, and overall training procedure unchanged.

\section{Experiments}
\label{sec:experiments}
\begin{table}[t]
\caption{\textbf{Training efficiency.} Wall-clock time to the best six-benchmark average achieved by OPDLM at each model scale. d-OPD consistently reaches the matched OPDLM performance faster, with speedups of $1.35$--$1.58\times$ across Qwen3 models from 0.6B to 8B.}
    \label{tab:time-to-target}
    \centering
\begin{tabular}{lcccc}
\toprule
Model scale & Qwen3-0.6B & Qwen3-1.7B & Qwen3-4B & Qwen3-8B \\
\midrule
OPDLM Time (hr)
& 18.39 & 19.92 & 30.72 & 28.26 \\

d-OPD Time (hr)
& 13.15 & 14.75 & 22.34 & 17.91 \\

\rowcolor{dopdbg}
\textbf{Speedup}
& \hspace{3.3pt}\textbf{1.40$\times$}
& \hspace{3.3pt}\textbf{1.35$\times$}
& \hspace{3.3pt}\textbf{1.38$\times$}
& \hspace{3.3pt}\textbf{1.58$\times$} \\
\bottomrule
\end{tabular}
\end{table}

\paragraph{Models.}
We evaluate Qwen3 checkpoints at 0.6B, 1.7B, 4B, and 8B parameters~\citep{Qwen32025Arxiv}. Model-scale comparisons use block size $N=4$, and we additionally train Qwen3-8B with $N=\{8,16\}$ to study larger blocks. Each pretrained AR checkpoint initializes the corresponding block dLLM student and remains frozen as both the causal teacher and future-compatibility scorer.

\paragraph{Training.}
We use the OPDLM training mixture~\citep{OPDLM2026Arxiv}, containing 61,816 prompts with maximum response length 4,000. Training uses one pass over the full prompt mixture for 483 on-policy iterations on 8 NVIDIA H100 GPUs. Within each comparison setting, methods use the same prompts, optimization budget, random seeds, and state-sampling procedure. Unless otherwise specified, d-OPD uses $k=16$. Full training and implementation details are provided in Appendix~\ref{app:reproducibility}.

\paragraph{Baselines.}
OPDLM~\citep{OPDLM2026Arxiv} is our primary baseline, which uses the same on-policy student states as d-OPD but directly distills the causal AR teacher distribution without future-aware correction. We additionally compare with SFT~\citep{SFT2025NIPS}, which applies standard masked-token supervised fine-tuning, Blockwise SFT (B-SFT)~\citep{BSFT2025Arxiv}, which adapts supervised fine-tuning to the blockwise generation structure, and BARD~\citep{BARD-VL2026Arxiv}, which converts AR models through progressive blockwise distillation. Together, these methods provide strong representative baselines for AR-to-block-dLLM conversion.

\paragraph{Evaluation.}
We evaluate MMLU~\citep{MMLU22021ICLR,MMLU12021ICLR}, MMLU-Pro~\citep{MMLU-Pro2024NIPS}, GPQA-Diamond~\citep{GPQA2024CoLM}, GSM8K~\citep{GSM8K2021Arxiv}, MATH500~\citep{MATH5002021NIPS}, and AIME25~\citep{AIME2025Arxiv}. Table~\ref{tab:main} reports six benchmark scores and their average at the checkpoint with the best six-benchmark average during training, following Appendix~\ref{app:reproducibility}.

\subsection{Main results}
Table~\ref{tab:main} summarizes the main results across model scales and block sizes. d-OPD consistently outperforms OPDLM across all evaluated Qwen3 scales, with gains of $2.6$--$3.9$ points in the six-benchmark average from 0.6B to 8B. The improvement remains stable over this wide range of model capacities, indicating that the benefit of future-aware correction is not specific to a particular student scale. On Qwen3-8B and Qwen3-4B, where we additionally compare against SFT, B-SFT, and BARD, d-OPD also achieves the strongest converted-model average. These results show that the gain is not limited to improving over the corresponding on-policy baseline, but remains competitive against alternative AR-to-block-dLLM conversion strategies.

The advantage also persists as the block size increases. On Qwen3-8B, d-OPD improves the six-benchmark average over OPDLM across all evaluated block sizes, with gains of up to $4.0$ points. The improvement remains substantial from block size $4$ to $16$, where the student can exploit more within-block future context and future-aware correction becomes increasingly important. This is consistent with the target-conditioning mismatch identified in Section~\ref{sec:motivation}: as more tokens within the active block are jointly involved in prediction, the student conditions on increasingly richer information than the causal AR teacher. The sustained gains across larger block sizes therefore further support incorporating visible future information into the teacher target.
\subsection{Training efficiency}

For each model scale, we use the best six-benchmark average achieved by OPDLM as the target and measure the first wall-clock time at which d-OPD reaches the same performance. This time-to-target metric complements the best achieved performance by measuring how quickly each method reaches a common performance level. By using wall-clock time, it also accounts for the additional teacher computation introduced by future scoring, rather than comparing methods only by the number of optimization steps. As shown in Table~\ref{tab:time-to-target}, d-OPD reaches the matched OPDLM performance $1.35$--$1.58\times$ faster across all four Qwen3 model scales, with the largest speedup on Qwen3-8B. The consistent improvement across scales indicates that the benefit is not limited to a particular model size. Thus, although future-aware scoring introduces additional per-step computation, the more informative teacher target enables the student to make faster progress toward the same performance level, yielding a net reduction in wall-clock training time.
\subsection{Ablation studies}
\label{subsec:ablation}

\begin{table}[t]
    \centering
    \caption{\textbf{Ablation study.} \textbf{Left:} Effect of candidate-set size $k$, showing the probability mass covered by the selected candidates and the resulting training overhead. Even small candidate sets cover nearly all teacher and student probability mass, while $k=16$ provides a favorable trade-off between coverage and overhead. \textbf{Right:} Effect of correctness gating. The gate improves the six-benchmark average from 58.8 to 59.8 by restricting future-aware correction to verifier-correct rollouts.}
    \begin{minipage}[t]{0.65\textwidth}
        \centering
        \begin{tabular}{lccccc}
            \toprule
            Mass (\%)
            & $k=4$
            & $k=8$
            & \cellcolor{dopdbg}\textbf{$k=16$}
            & $k=32$
            & $k=64$ \\
            \midrule
            Teacher
            & 97.87
            & 98.88
            & \cellcolor{dopdbg}\textbf{99.45}
            & 99.67
            & 99.79 \\
            Student
            & 95.43
            & 98.30
            & \cellcolor{dopdbg}\textbf{99.44}
            & 99.64
            & 99.77 \\
            Overhead
            & 2.44\%
            & 2.57\%
            & \cellcolor{dopdbg}\textbf{2.63\%}
            & 3.50\%
            & 6.02\% \\
            \bottomrule
        \end{tabular}
    \end{minipage}
    \hfill
    \begin{minipage}[t]{0.31\textwidth}
        \centering
        \begin{tabular}{lcc}
            \toprule
            Method & Gate & Avg. \\
            \midrule
            OPDLM & -- & 55.9 \\
            d-OPD & Off & 58.8 \\
            \cellcolor{dopdbg}\textbf{d-OPD}
            & \cellcolor{dopdbg}\textbf{On}
            & \cellcolor{dopdbg}\textbf{59.8} \\
            \bottomrule
        \end{tabular}
    \end{minipage}
    \label{tab:ablation}
\end{table}
Table~\ref{tab:ablation} first examines candidate restriction. Even very small candidate sets capture most of the probability mass under both the teacher and student distributions, and the coverage quickly saturates as $k$ increases. By $k=16$, the candidate set already covers more than $99\%$ of the probability mass for both distributions, while the measured training overhead remains only $2.63\%$. Increasing $k$ further yields only marginal gains in coverage but noticeably higher overhead, indicating that most of the useful future correction is concentrated on a small number of high-probability tokens. This also suggests that full-vocabulary future scoring is unnecessary in practice: restricting the correction to likely candidates preserves nearly all of the probability mass relevant to the teacher target while substantially reducing computation. We therefore use $k=16$ as the default coverage--cost tradeoff.

The right side of Table~\ref{tab:ablation} evaluates correctness gating. Applying future-aware correction only to verifier-correct rollouts improves the six-benchmark average from $58.8$ to $59.8$. This suggests that completed futures from successful trajectories provide more reliable evidence for reweighting, while the gate also avoids unnecessary future scoring on incorrect rollouts and reduces training cost.
\section{Conclusion}
On-policy distillation for AR-to-block-dLLM conversion supervises student-visited states with a causal AR teacher. We identify a target-conditioning mismatch in this setting: the standard teacher target depends only on the causal prefix, while the block dLLM student uses the full visible state of the active block, including visible future context. We introduce d-OPD, which uses visible future context within each block to reweight the causal AR teacher distribution toward a target that better matches the information available to the student. Across Qwen3 models from 0.6B to 8B, d-OPD consistently improves over OPDLM, with gains of up to $4.0$ points in the six-benchmark average and up to $1.58\times$ speedup to matched OPDLM performance. These results highlight the effectiveness of correcting the teacher target with student-visible future context.

\subsection*{AI use statement}

In this work, we used generative AI tools for assisting with method implementation. We have not used generative AI tools for research ideation, methodology or experiment design, hypothesis development, mathematical claims or proofs, data analysis, or result interpretation, and the rest of the required disclosure tasks are not applicable to this work. Additionally, we used generative AI tools for language polishing, assisting with evaluation and experiment code, and monitoring experiment execution. We have reviewed all AI-assisted work. All AI-assisted code was manually reviewed, tested, and verified by the authors, and all AI-assisted text was manually checked for technical accuracy and consistency with the paper. We take responsibility for the final content of this work, including text, claims or artifacts produced with the aid of generative AI.

\subsection*{Ethics statement}

This work studies training methods for language models and does not involve human subjects or the collection of private or personally identifiable information. Our experiments use existing models, benchmarks, and training data following their intended research use. We are not aware of additional ethical concerns specific to the proposed method beyond the broader considerations associated with training and deploying large language models in practical settings.

\subsection*{Reproducibility statement}

We provide the main algorithm and training objective in Section~\ref{sec:method}, with the detailed training procedure summarized in Algorithm~\ref{alg:training}. Appendix~\ref{app:reproducibility} documents the training setup, rollout and state-sampling procedure, correctness gating, benchmark evaluation, and checkpoint-selection protocol used throughout the experiments. Appendix~\ref{app:enumerable} gives the complete construction of the controlled enumerable setting used for the exact-target diagnosis, including the restricted support, compared targets, and KL aggregation procedure. Appendix~\ref{app:timing} describes the wall-clock timing protocol used to measure both time to matched performance and the additional training overhead introduced by future-aware scoring. Additional implementation details, including candidate scoring and caching, are provided in Appendix~\ref{app:implementation}, while Appendix~\ref{app:ablations} reports the full correctness-gating ablation results. Together, these sections specify the main algorithmic, training, evaluation, diagnostic, and timing choices needed to faithfully reproduce the reported results and comparisons.

\bibliography{iclr2027_conference}
\bibliographystyle{iclr2027_conference}

\appendix
\section{Training Algorithm}
\label{app:algorithm}

Algorithm~\ref{alg:training} summarizes one on-policy training iteration of d-OPD. Given student-visited denoising states and the completed trajectories produced by the same rollout, d-OPD constructs a future-aware teacher target for each masked prediction position during student training. The correction is applied only to a small candidate set and only when the completed rollout passes the task verifier. Otherwise, training falls back to the ordinary causal teacher target used for standard distillation.

\begin{algorithm}[t]
    \caption{One on-policy training iteration of d-OPD}
    \label{alg:training}
    \begin{algorithmic}[1]
        \Require Student $p_\theta$, frozen AR teacher $p_T$, candidate size $k$
        \State Generate responses with $p_\theta$ and record visited denoising states
        \State Obtain each completed response $\hat{y}$ and its correctness label $R(\hat{y})$
        \State Sample one visited denoising state for each generated block
        \For{each selected active-block state $X_t$}
            \State Let $z$ be the completed causal context preceding the active block
            \State Let $\hat{X}=(\hat{x}_1,\ldots,\hat{x}_N)$ be its completed block from the same rollout
            \State Run $p_\theta$ on $X_t$ to obtain $p_{\theta,i}$ for all masked positions $i$
            \For{each masked position $i$ in $X_t$}
                \State Form completed prefix $\hat{A}_i=(\hat{x}_1,\ldots,\hat{x}_{i-1})$
                \State Compute causal prior $r_i(v)=p_T(v\mid q,z,\hat{A}_i)$
                \If{$R(\hat{y})=1$ and $X_t$ contains visible future context for $i$}
                    \State Construct $V_i^{(k)}$ as the union of the teacher and student top-$k$ sets
                    \State Form completed suffix $\hat{C}_i=(\hat{x}_{i+1},\ldots,\hat{x}_N)$
                    \State Compute reference score $s_i(\hat{x}_i)=\log p_T(\hat{C}_i\mid q,z,\hat{A}_i,\hat{x}_i)$
                    \For{$v\in V_i^{(k)}$}
                        \State Compute $s_i(v)=\log p_T(\hat{C}_i\mid q,z,\hat{A}_i,v)$
                        \State Compute $\Delta s_i(v)=s_i(v)-s_i(\hat{x}_i)$
                    \EndFor
                    \State Reweight candidates in $V_i^{(k)}$ by $\exp(\Delta s_i(v))$ and normalize to obtain $\tilde p_{T,i}$
                \Else
                    \State Set $\tilde p_{T,i}=r_i$
                \EndIf
            \EndFor
        \EndFor
        \State Average the token-level losses over all masked positions in each rollout
        \State Update $\theta$ using Equation~\ref{eq:training-loss}
    \end{algorithmic}
\end{algorithm}

\section{Additional Implementation Notes}
\label{app:implementation}

\paragraph{Candidate scoring and caching.}
For each masked position eligible for future-aware correction, we score only the union of the teacher and student top-$k$ sets. The teacher prefix $(q,z,\hat{A}_i)$ is shared across all candidate branches for a given position and can be cached and reused. Only the candidate token differs across branches, and the same completed suffix $\hat C_i$ is then scored under each candidate. This avoids recomputing the shared causal prefix for every candidate and keeps teacher computation focused on the part of the sequence that depends on the candidate choice. Tokens outside the candidate set are left unchanged by the future-aware correction before the final normalization, so the sparse scoring procedure modifies only the selected high-probability candidates.

For each eligible position, we construct one scoring job containing a reference branch for the realized rollout token $\hat{x}_i$ and one branch for every distinct token in the union of the teacher and student top-$k$ sets. With the default $k=16$, each job therefore contains at most 33 branches, including the reference branch. All branches for the same position are evaluated together in a single scorer batch in the default implementation. This organization allows most of the shared causal-prefix computation to be reused rather than recomputed independently for every candidate.

For each training example, we run the frozen AR scorer once on the clean prompt--response sequence to construct a Hugging Face \texttt{DynamicCache}~\citep{Transformers2020ACL}. For an individual scoring job, this cache is truncated to the causal prefix preceding the candidate position. At every Transformer layer, the cached key and value tensors have logical shape $[1,n_{\mathrm{kv}},L_{\mathrm{prefix}},d_{\mathrm{head}}]$ and are expanded only along the candidate-branch batch dimension. The suffix beginning at the candidate position is then evaluated separately for each branch. The same trajectory-level cache is reused across all eligible positions in that training example, but caches are not shared across different examples. To avoid excessive memory growth from cache expansion, the candidate-branch batch is further reduced whenever the estimated expanded prefix-cache size would exceed 16\,GiB.

\paragraph{Fallback behavior and numerical safeguards.}
The corrected target reduces exactly to the causal teacher target whenever the response fails the correctness gate or when no future token is visible to the right of the masked position in the sampled student state. In both cases, there is no future-aware reweighting and we directly set $\tilde p_{T,i}=r_i$.

The scorer is loaded in the checkpoint's native dtype, which is bfloat16 for the reported Qwen3 experiments. Scorer logits, teacher and student log probabilities, suffix log-likelihood sums, relative score differences, correction factors, and final normalization are computed in float32. Candidate scores are expressed relative to the reference score $s_i(\hat{x}_i)$ before exponentiation. The future-aware correction for a position is committed only when the resulting full-vocabulary normalizer is finite and strictly positive. Otherwise, the causal target $r_i$ is retained for that position. We do not otherwise clip finite score differences. Exact zero-probability entries remain zero after correction and are represented as $-\infty$ when converted back to log probabilities.

\section{Training and Evaluation Details}
\label{app:reproducibility}

\subsection{Data}

We use the OPDLM training mixture~\citep{OPDLM2026Arxiv}, which contains 61,816 prompts in total. We follow the same data preprocessing and prompt formatting used by OPDLM unless otherwise specified. The original task information required for training and verification is retained throughout preprocessing. Each prompt is rendered with the Qwen chat template using \texttt{add\_generation\_prompt=True} and \texttt{enable\_thinking=False}. We perform no additional deduplication or filtering beyond the preprocessing inherited from the original OPDLM setup.
\subsection{Training}

Within each comparison setting, methods use the same prompt mixture, rollout configuration, state-selection procedure, optimization budget, and random seeds. Training makes one pass over the full prompt mixture, corresponding to 483 on-policy iterations on eight NVIDIA H100 80\,GB GPUs. Unless otherwise specified, d-OPD uses candidate size $k=16$.

Each on-policy iteration processes up to 128 prompts and generates one rollout per prompt with temperature $1.0$, top-$p=1.0$, and top-$k=0$. Rollout decoding uses four denoising steps per block and a maximum generated response length of 4,000 tokens. At on-policy iteration $s$, a global response-length limit $M_s$ increases from 100 to 4,000 tokens over the first 100 iterations according to a cosine schedule. We otherwise follow the rollout configuration used by OPDLM.

For state selection, we record the first-unmask time of every generated token and reconstruct the denoising states actually visited during rollout. For each response block, we uniformly sample one visited denoising round in which that block contains at least one masked prediction position. The round is sampled independently across blocks, and the selected blockwise states are then combined into one trajectory-level training example. Different blocks within the same example may therefore correspond to different denoising rounds, but every selected state was visited during the same on-policy rollout. Completed preceding blocks provide the corresponding causal context $z$, while the final completed rollout provides $\hat{X}$, $\hat{A}_i$, and $\hat{C}_i$ used for teacher-target construction.

We optimize the student with AdamW~\citep{AdamW2019ICLR} using learning rate $1\times10^{-5}$, $\beta_1=0.9$, $\beta_2=0.999$, $\epsilon=10^{-8}$, and zero weight decay. The learning rate is linearly warmed up during the first five on-policy iterations and then cosine-decayed to $10\%$ of its peak value. Gradients are clipped to a global norm of $1.0$. We use BF16 mixed precision, TF32 matrix operations, gradient checkpointing, and DeepSpeed ZeRO Stage~3~\citep{ZeRO2020C,DeepSpeed2020KDD,ZeRO-Offload2021ATC} with parameters and optimizer states offloaded to CPU.

Each on-policy iteration generates up to 128 rollouts, one for each sampled prompt. Every retained rollout yields one trajectory-level training example containing the sampled blockwise denoising states described above. These examples are then used for student optimization under the training configuration described above. Keeping each rollout as a separate training example preserves the direct correspondence between its visited states and final completed trajectory throughout target construction. We do not pack multiple trajectory-level examples into a single sequence.

\subsection{Verifier and correctness gating}

We follow the task-specific verification procedure used in the OPDLM training setup and use the resulting Boolean correctness signal as $R(\hat{y})$. Verification is applied only to determine whether the future-aware correction is enabled for a rollout. A verifier-correct rollout receives the future-aware teacher correction described in Section~\ref{sec:method}, whereas a verifier-incorrect rollout continues to use the standard causal teacher target. Thus, correctness gating does not discard unsuccessful trajectories from training and affects only the additional future-aware reweighting. Examples without an applicable correctness verifier receive standard causal teacher supervision.

\subsection{Benchmark evaluation}

We evaluate MMLU, MMLU-Pro, GPQA-Diamond, GSM8K, MATH500, and AIME25 using the repository-local OPDLM evaluation pipeline from our released code snapshot~\citep{OPDLM2026Arxiv}. We follow the corresponding OPDLM benchmark formatting, answer extraction, and verification procedures. All six benchmarks are evaluated zero-shot using the Qwen3 chat template.

Evaluation uses the same block length as the corresponding model setting, four denoising steps per block, low-confidence static remasking, temperature $1.0$, top-$p=1.0$, and top-$k=1$. We generate one completion per problem with at most 4,096 new tokens. Evaluation is performed on the full benchmark sets: 14,042 MMLU examples, 12,032 MMLU-Pro examples, 198 GPQA-Diamond examples, 1,319 GSM8K examples, 500 MATH500 examples, and 30 AIME25 examples.

We evaluate checkpoints after iterations 50, 100, \ldots, 450 and once more at the final iteration 483. For each evaluated checkpoint, we compute the unweighted average accuracy over the six benchmarks. We then select the checkpoint with the highest six-benchmark average and report all six benchmark scores from this same checkpoint. The reported aggregate is the corresponding six-benchmark average. We do not independently select the best checkpoint for each benchmark.

\section{Enumerable Teacher Construction and Exact-Target Diagnosis}
\label{app:enumerable}

We construct a finite teacher distribution induced by Qwen3-8B so that the exact future-conditioned target can be evaluated by exhaustive enumeration. The purpose of this experiment is diagnostic: it isolates the target-conditioning mismatch in a controlled setting where the complete teacher-induced posterior can be computed exactly, allowing candidate teacher targets to be compared directly against the exact posterior rather than only through downstream task accuracy.

\subsection{Restricted enumerable support}

We construct a restricted enumerable vocabulary sequentially from left to right under a fixed causal context. Suppose the first $j$ positions have already been assigned restricted vocabularies $\mathcal{S}_1,\mathcal{S}_2,\ldots,\mathcal{S}_j$ of size $V$, defining a set $\mathcal{A}_j$ of $V^j$ possible restricted prefixes. For each prefix $A\in\mathcal{A}_j$, we query Qwen3-8B for the full next-token distribution and assign the prefix the normalized teacher weight
\begin{equation}
\omega(A)=
\frac{
p_T(A\mid q)
}{
\sum_{A'\in\mathcal{A}_j}p_T(A'\mid q)
}.
\end{equation}
We then form the probability-weighted mixture
\begin{equation}
\bar p_{j+1}(v)=
\sum_{A\in\mathcal{A}_j}
\omega(A)p_T(v\mid q,A).
\end{equation}
The top-$V$ tokens under $\bar p_{j+1}$ define the restricted vocabulary $\mathcal{S}_{j+1}$ at position $j+1$. Repeating this procedure yields a finite support that preserves high-probability local choices under the original AR teacher while remaining small enough for exhaustive enumeration.

\subsection{Exact teacher posterior}

After constructing the restricted vocabulary at every position, we enumerate every complete block sequence $X$ in the resulting finite support. Each sequence is scored using the original AR factorization,
\begin{equation}
p_T(X\mid q)=
\prod_{j=1}^{N}
p_T(x_j\mid q,x_{<j}),
\end{equation}
and the resulting probabilities are renormalized over the restricted support. For a partially observed state $X_t$, exhaustive enumeration then gives the exact block posterior in Equation~\ref{eq:block-posterior} and the corresponding exact token marginal in Equation~\ref{eq:token-posterior}. Because every compatible completion is explicitly enumerated, this construction provides a direct reference target against which approximate teacher targets can be measured consistently under the same restricted support.
\subsection{Compared targets}

For each partially observed state $X_t$, we sample one compatible completion
\begin{equation}
\hat X=(\hat x_1,\ldots,\hat x_N)
\end{equation}
from the corresponding completion distribution. For each masked position $i$, this completion provides the completed prefix $\hat A_i=\hat x_{<i}$ and completed suffix
\begin{equation}
\hat C_i=(\hat x_{i+1},\ldots,\hat x_N).
\end{equation}

The causal target is
\begin{equation}
r_i(v)=p_T(v\mid q,\hat A_i),
\end{equation}
matching the ordinary causal AR supervision used by OPDLM. The future-aware target additionally scores how compatible the same completed future is with each candidate token,
\begin{equation}
s_i(v)=
\log p_T(\hat C_i\mid q,\hat A_i,v),
\end{equation}
and forms the relative score
\begin{equation}
\Delta s_i(v)=s_i(v)-s_i(\hat x_i).
\end{equation}
The resulting future-aware target is obtained by reweighting the causal teacher distribution with $\exp(\Delta s_i(v))$ and normalizing over the restricted support. Because the enumerable support is small, this diagnostic applies the correction to the full restricted vocabulary rather than introducing the practical top-$k$ approximation used in large-scale training. This allows the experiment to focus directly on whether incorporating completed future information moves the causal teacher target closer toward the exact future-conditioned posterior under controlled conditions.

\subsection{KL metric and aggregation}

For each masked position $i$, we measure target quality using
\begin{equation}
D_{\mathrm{KL}}
\left(
\pi_i(\cdot\mid X_t,q)
\mathrel{\|}
\hat p_i(\cdot\mid X_t,q)
\right),
\end{equation}
where $\pi_i$ is the exactly enumerated future-conditioned target from Equation~\ref{eq:token-posterior}, and $\hat p_i$ denotes either the causal or future-aware target defined above. The KL divergence is computed over the restricted candidate support at position $i$, providing a consistent token-level measure of target quality.

We evaluate block lengths $N\in\{4,6,8\}$ and restrict each position to $V$ candidate tokens drawn from its position-specific enumerable support. This gives the ten configurations
\begin{equation}
(N,V)\in
\{
(4,6),(4,12),(4,24),(4,48),
(6,6),(6,8),(6,12),(6,16),
(8,6),(8,8)
\}.
\end{equation}
For each configuration, we sample 400 complete blocks from the restricted teacher distribution and independently retain each position with probability $0.45$, masking the remaining positions. If all positions are retained, we uniformly select one position to mask so that every sampled state contains at least one prediction target. The exact target for every masked position is then obtained by exhaustive enumeration over the corresponding finite support.

To vary the discrepancy between the student and teacher completion distributions, we define a perturbed student distribution over the same finite complete-block support:
\begin{equation}
p_S^{(\sigma)}(X)
\propto
\exp\left(
\log p_T(X)+\sigma Z_X
\right),
\qquad
Z_X\overset{\mathrm{i.i.d.}}{\sim}\mathcal N(0,1),
\end{equation}
with $\sigma\in\{0,0.5,1.0,1.5\}$. At $\sigma=0$, the student distribution coincides with the restricted teacher distribution, while larger $\sigma$ introduces stronger perturbations. The same 400 partially observed states are reused across all four noise levels. For each state and noise level, we restrict $p_S^{(\sigma)}$ to compatible complete blocks, renormalize it, and sample one completion $\hat X$ used to construct the causal and future-aware targets above, enabling a controlled comparison across perturbation strengths.

When no future token is visible to the right of position $i$, the exact future-compatibility term is constant across candidates, so the future-aware target reduces to the causal teacher target. These positions therefore provide a natural limiting case in which the two compared targets should coincide.

Across all configurations, noise levels, sampled states, and masked positions, the sweep contains 49,948 masked-token marginals. We pool the token-level KL values over the full sweep and report their arithmetic mean as the aggregate diagnostic statistic used in our analysis.

\begin{table}[t]
    \caption{
    Mean KL divergence to the exactly enumerated future-conditioned target, averaged over 49,948 masked-token marginals across all controlled configurations. Lower is better.
    }
    \label{tab:diagnosis}
    \centering
    \begin{tabular}{lcc}
        \toprule
        Target & KL & Reduction \\
        \midrule

        Causal teacher
        & 0.4444
        & -- \\

        \cellcolor{dopdbg}\textbf{Future-aware}
        & \cellcolor{dopdbg}\textbf{0.1249}
        & \cellcolor{dopdbg}\textbf{71.9\%} \\

        \bottomrule
    \end{tabular}
\end{table}

\section{Wall-Clock Timing Protocol}
\label{app:timing}

Training efficiency is measured using time to matched performance. For each model scale, we first identify the best six-benchmark average achieved by OPDLM under the checkpoint-selection protocol described in Appendix~\ref{app:reproducibility}. This value is then used as the matched-performance target for the corresponding comparison. We measure the elapsed wall-clock time at which OPDLM reaches this value and the first elapsed time at which d-OPD reaches or exceeds the same threshold.

We separately measure the wall-clock overhead introduced by future-aware scoring during training for Qwen3-0.6B, 1.7B, 4B, and 8B. Each measured duration covers the complete on-policy iteration and therefore includes rollout generation, verifier computation, teacher scoring, student optimization, model serialization, and reloading the updated student weights into the rollout engine. Periodic benchmark evaluation is performed outside the timed region and is excluded from the measurement, while routine training and checkpointing costs remain included so that the reported timing reflects the actual end-to-end training workflow under the same experimental setup as closely as possible.

All measurements use eight NVIDIA H100 80\,GB GPUs and the same distributed configuration for OPDLM and d-OPD at each model scale. For each method and model scale, we compute the mean wall-clock duration per on-policy iteration and use this quantity to compare their per-iteration training cost. The relative overhead is defined as the percentage increase in mean iteration time of d-OPD over OPDLM, and we report its unweighted average across the four evaluated model scales. Under the default $k=16$ configuration, this yields the $2.63\%$ average per-iteration overhead reported in the main text. Candidate restriction and correctness gating are both enabled in this measurement, matching the default d-OPD training configuration used in the main experiments.

\section{Limitations and Future Work}
\label{app:limitations}

d-OPD currently uses a single completed on-policy trajectory to construct the future-aware target. A natural extension is to sample multiple compatible trajectories for the same student-visited state and aggregate their future evidence when reweighting the teacher distribution. Averaging across several possible futures may reduce the variance introduced by relying on a single sampled completion and provide a more stable estimate of the future-conditioned teacher target. In addition, d-OPD is complementary to supervised fine-tuning because it modifies the teacher target during on-policy distillation rather than the student architecture or the underlying training objective. This makes it straightforward to combine the two paradigms in a two-stage training pipeline, using SFT for initial adaptation followed by d-OPD for on-policy refinement. Studying how these two stages interact, and whether SFT provides a stronger initialization for subsequent future-aware distillation, is an interesting direction for future work. More broadly, evaluating d-OPD across additional model families and block sizes is also an important direction for future work.

\section{Additional Ablation Results}
\label{app:ablations}

Table~\ref{tab:gate-full} provides the full per-benchmark results for the correctness gate ablation reported in Table~\ref{tab:ablation}. 

\begin{table}[H]
    \caption{
    Full results for the correctness-gating ablation on Qwen3-8B. The gate applies future-aware correction only to verifier-correct rollouts, while ordinary causal teacher supervision remains active for all rollouts. Avg. denotes the unweighted average over the six benchmarks. Higher is better.
    }
    \label{tab:gate-full}
    \centering
    \setlength{\tabcolsep}{5.5pt}
    \begin{tabular}{llccccccc}
        \toprule
        Method & Gate
        & MMLU & MMLU-P & GPQA-D
        & GSM8K & MATH500 & AIME25
        & Avg. \\
        \midrule

        OPDLM
        & --
        & 69.2
        & 54.0
        & 38.4
        & 87.2
        & 76.8
        & 10.0
        & 55.9 \\

        d-OPD
        & Off
        & 74.2
        & \textbf{56.9}
        & 37.4
        & 87.9
        & 76.2
        & 20.0
        & 58.8 \\

        \rowcolor{dopdbg}
        \textbf{d-OPD}
        & \textbf{On}
        & \textbf{74.3}
        & 55.9
        & \textbf{38.9}
        & \textbf{89.3}
        & \textbf{77.2}
        & \textbf{23.3}
        & \textbf{59.8} \\

        \bottomrule
    \end{tabular}
\end{table}

\end{document}